\documentclass[conference]{IEEEtran}
\IEEEoverridecommandlockouts
\usepackage{cite}
\usepackage{amsmath,amssymb,amsfonts}
\usepackage{algorithm}
\usepackage{pifont}
\newcommand{\cmark}{\ding{51}} 
\newcommand{\xmark}{\ding{55}} 
\usepackage{graphicx}
\usepackage{textcomp}
\usepackage{xcolor}
\usepackage{float}
\usepackage[colorlinks=true, urlcolor=blue]{hyperref}
\usepackage{algpseudocode} 
\usepackage{url}           

\usepackage{listings} 

\lstdefinestyle{json}{
    basicstyle=\ttfamily\scriptsize,
    breaklines=true,
    postbreak=\mbox{\textcolor{red}{$\hookrightarrow$}\space},
}
\def\BibTeX{{\rm B\kern-.05em{\sc i\kern-.025em b}\kern-.08em
    T\kern-.1667em\lower.7ex\hbox{E}\kern-.125emX}}

\begin{document}

\title{XMorph: Explainable Brain Tumor Analysis \\Via LLM-Assisted Hybrid Deep Intelligence}

\author{
\IEEEauthorblockN{Sepehr Salem Ghahfarokhi\textsuperscript{1}, M.~Moein Esfahani\textsuperscript{2}, Raj Sunderraman\textsuperscript{1}, Vince Calhoun\textsuperscript{2}, Mohammed Alser\textsuperscript{1}}
\vspace{0.15cm} \\
\IEEEauthorblockA{
\textsuperscript{1}\textit{Department of Computer Science, Georgia State University}, Atlanta, GA, USA \\
\textsuperscript{2}\textit{TReNDS Center, Georgia State University}, Atlanta, GA, USA \\
Corresponding authors: ssalemghahfarokhi1@gsu.edu, malser@gsu.edu
}
}

\maketitle
\thispagestyle{plain}
\pagestyle{plain}

\begin{abstract}
Deep learning has significantly advanced automated brain tumor diagnosis, yet clinical adoption remains limited by interpretability and computational constraints. Conventional models often act as opaque ``black boxes'' and fail to quantify the complex, irregular tumor boundaries that characterize malignant growth. To address these challenges, we present XMorph, an explainable and computationally efficient framework for fine-grained classification of three prominent brain tumor types: glioma, meningioma, and pituitary tumors. We propose an Information-Weighted Boundary Normalization (IWBN) mechanism that emphasizes diagnostically relevant boundary regions alongside nonlinear chaotic and clinically validated features, enabling a richer morphological representation of tumor growth. A dual-channel explainable AI module combines GradCAM++ visual cues with LLM-generated textual rationales, translating model reasoning into clinically interpretable insights. The proposed framework achieves a classification accuracy of 96.0\%, demonstrating that explainability and high performance can co-exist in AI-based medical imaging systems.
The source code and materials for XMorph  are all publicly available at: \url{https://github.com/ALSER-Lab/XMorph}.
\end{abstract}


\section{Introduction}
Brain tumors are among the most lethal forms of cancer, with patient prognosis heavily dependent on early and accurate diagnosis. Magnetic Resonance Imaging (MRI) is the standard modality for non-invasive tumor detection, and the rise of deep learning, particularly Convolutional Neural Networks (CNNs), has led to significant advances in automated image analysis \cite{b1}. These methods have been especially effective in classifying \emph{gliomas}, \emph{meningiomas}, and \emph{pituitary tumors}—three of the most prevalent and clinically consequential intracranial tumor types. Together, these tumors account for the majority of adult brain tumor diagnoses and present distinct morphological and clinical characteristics, making them ideal targets for developing explainable and fine-grained AI-based diagnostic systems.

However, three major challenges hinder the widespread clinical adoption of these systems. 

First, the inherent ``black box'' nature of deep neural networks creates a substantial trust deficit among clinicians \cite{b2}. Without transparent insight into a model’s reasoning, its predictions are difficult to rely on for critical patient-care decisions, a limitation widely discussed in medical AI surveys \cite{b3}.

Second, high-performing models are often computationally expensive. For example, the MGMT-net model reports 97.2\% accuracy but requires 3.42 seconds per MRI slice during inference due to its deep architecture \cite{b4}. Such heavy networks impose significant memory and processing demands, reducing their practicality in real-time diagnostic workflows and resource-limited clinical settings—an issue further magnified by the rise of portable MRI devices.

Third, existing approaches often struggle to capture the irregular, non-Euclidean tumor boundaries characteristic of malignant growth. Building on our previous work \cite{b6,b10}, we emphasize the importance of metrics from nonlinear dynamics—such as Fractal Dimension (FD) and Approximate Entropy (ApEn)—which are computationally lightweight yet sensitive to subtle boundary irregularities frequently overlooked by CNN-based features.

To address the first two challenges, recent studies have explored two primary directions. Post-hoc explainability tools, such as GradCAM++, are widely used to highlight influential regions in MRI slices \cite{b7,b9}, while LLM-based approaches generate textual rationales for model predictions \cite{b8}. In parallel, lightweight architectures aim to reduce parameter counts while maintaining competitive accuracy \cite{b9}. 

However, these solutions are typically developed in isolation and, critically, do not address the challenge of modeling complex tumor morphology through integrated clinical and mathematical descriptors. 

Our goal is to develop a diagnostic system that bridges the gap between high-performance automated classification and clinical transparency. To this end, we introduce XMorph, a clinically interpretable framework for the fine-grained classification of glioma, meningioma, and pituitary tumors. The key idea of XMorph is the synergistic integration of deep convolutional features, metrics from nonlinear dynamics, and quantitative clinical biomarkers into a unified diagnostic pipeline. To our knowledge, XMorph is the first framework to provide a dual-channel explainability module that interprets both handcrafted morphological features and radiological markers through natural language, grounding AI predictions in established clinical practice. We evaluate XMorph through a comparative analysis against prominent state-of-the-art methodologies, demonstrating that our framework provides a superior foundation for interpretable brain tumor analysis.

\noindent The contributions of this work are as follows:
\begin{itemize}
    \item \textbf{Novel Information-Weighted Boundary Normalization (IWBN):} a first-of-its-kind boundary enhancement method that automatically prioritizes informative boundary segments, producing a more discriminative characterization of infiltrative growth than conventional global shape features.

    \item \textbf{Hybrid Feature Representation:} a unified feature space that fuses CNN embeddings, metrics from nonlinear dynamics (FD, ApEn, LE), and clinical biomarkers (REI, MLS), enabling richer and more stable tumor characterization.

    \item \textbf{Dual-channel Explainable AI (XAI):} a modular XAI system that couples spatial visualizations (GradCAM++) with LLM-generated clinical narratives, effectively translating the reasoning behind our entire hybrid feature set into human-readable insights.

    \item \textbf{Comprehensive Benchmarking and Efficiency:} We demonstrate through quantitative benchmarks that XMorph achieves competitive classification accuracy while operating with significantly lower computational overhead than heavy state-of-the-art architectures.
\end{itemize}

\section{Related Work}
To our knowledge, XMorph is the first hybrid intelligence framework for brain tumor analysis that holistically integrates deep convolutional features, nonlinear boundary dynamics, and quantitative clinical biomarkers within a dual-channel explainable AI (XAI) module. While prior works have separately explored these areas, they often focus on either deep learning accuracy, post-hoc explainability, or nonlinear feature analysis in isolation. Our work bridges the gaps between these domains by proposing a unified, lightweight, and interpretable system. In this section, we review the existing literature across three key areas that our framework synthesizes: (1) CNN-based and nonlinear feature approaches for tumor analysis, (2) explainable AI in medical imaging, and (3) the emerging role of Large Language Models (LLMs).
\\
\subsubsection{CNN-Based and Nonlinear Feature Approaches for Brain Tumor Analysis} CNN architectures like VGG, DenseNet, and ResNet, often combined with transfer learning, are the cornerstone of modern image-based classification \cite{b1}. For precise localization, segmentation models such as YOLOv8 and DeepLabV3+ are commonly employed as a preliminary step to isolate the tumor region before classification \cite{b1,b2}. More recently, custom lightweight CNNs have been explored to reduce computational overhead. For instance, the work in \cite{b5} introduces a lightweight CNN optimized via a novel Nonlinear Lévy Chaotic Moth Flame Optimiser (NLCMFO) for efficient hyperparameter tuning, achieving high accuracy in an end-to-end fashion. Similarly, \cite{b11} proposes CDCG-UNet, a segmentation framework enhanced by Chaotic Harris Shrinking Spiral Optimization (CHSOA), which significantly boosts tumor boundary delineation on BRATS datasets. Another related effort \cite{b12} combines fuzzy logic with CNN and U-Net architectures to improve tumor classification and segmentation accuracy through nonlinear feature handling. While effective, these approaches primarily focus on architectural optimization and may not explicitly capture the rich information contained in the nonlinear dynamics of tumor growth. Furthermore, like many deep learning systems, they often lack built-in interpretability, a well-documented challenge that can hinder clinical trust and adoption \cite{b2, b3}.

Beyond chaos-inspired methods, several studies have explored nonlinear and fractal-based features for brain tumor analysis. The PFA-Net framework \cite{b13} embeds fractal dimension estimation into a CNN segmentation pipeline, effectively modeling the morphological complexity of tumor regions. The MFDNN model \cite{b14} incorporates multiresolution fractal texture features, derived from fractional Brownian motion, into deep networks for robust and uncertainty-aware tumor segmentation. In addition, \cite{b15} demonstrates that fractal dimension metrics of tumor structure, boundary, and skeleton are effective biomarkers for distinguishing high-grade from low-grade gliomas, achieving strong diagnostic accuracy. While these approaches enrich tumor representation with nonlinear morphological insights, they often lack integration with clinical priors or multimodal interpretability mechanisms.

\subsubsection{Explainable AI (XAI) in Medical Imaging} Recognizing the "black box" problem, researchers have begun integrating XAI techniques. Visualization methods like Gradient-weighted Class Activation Mapping (GradCAM) are widely used to produce heatmaps that highlight the image regions most influential to a model's decision\cite{b2,b7}. This provides a crucial visual aid for clinicians. However, these visual maps alone may not fully articulate the reasoning behind a classification, such as the specific textural or morphological properties the model identified.

Beyond GradCAM heatmaps, recent works have introduced alternative XAI strategies in medical imaging. Muhammad et al. \cite{b16} present a systematic review covering gradient-based, perturbation-based, and example-based methods, highlighting their applicability and limitations for clinical adoption. Dravid et al. \cite{b17} propose medXGAN, a GAN-based framework that perturbs input images in a learned manner to generate realistic counterfactuals, revealing which structures most influence a model’s decision and offering richer interpretability than saliency maps alone.

\subsubsection{The Role of Large Language Models (LLMs)} The advent of powerful LLMs has opened new frontiers for medical AI. Initial applications focused on generating diagnostic reports from model outputs, as seen with BioMistral-7B \cite{b8}, or using multimodal LLMs like LLaVA and GPT-4 for direct image classification via prompting \cite{b5,b9}. These models can provide ``soft explainability'' by generating textual descriptions of their findings. However, when used in isolation for classification, they may not leverage the specialized, fine-grained features that CNNs excel at extracting. Furthermore, work like EEG-GPT \cite{b3} shows that LLMs can interpret pre-extracted, Tumor Specific features, but this has not yet been explored in the context of chaotic features for brain tumor imaging.

 Despite these advances, a clear research gap remains. Existing models often: (a) provide either visual (GradCAM) or textual (LLM) explanations, but rarely integrate both into a cohesive framework; (b) rely on computationally intensive architectures that are not optimized for rapid, lightweight analysis; and (c) have not explored the rich, descriptive power of nonlinear chaotic features for characterizing tumor morphology. Our work directly addresses this gap by proposing a hybrid framework that is lightweight, multimodally explainable, and specifically tailored for the fine-grained, three-class brain tumor classification task.

\section{Proposed Methodology}

\begin{figure*}[h!]
  \includegraphics[width=\textwidth]{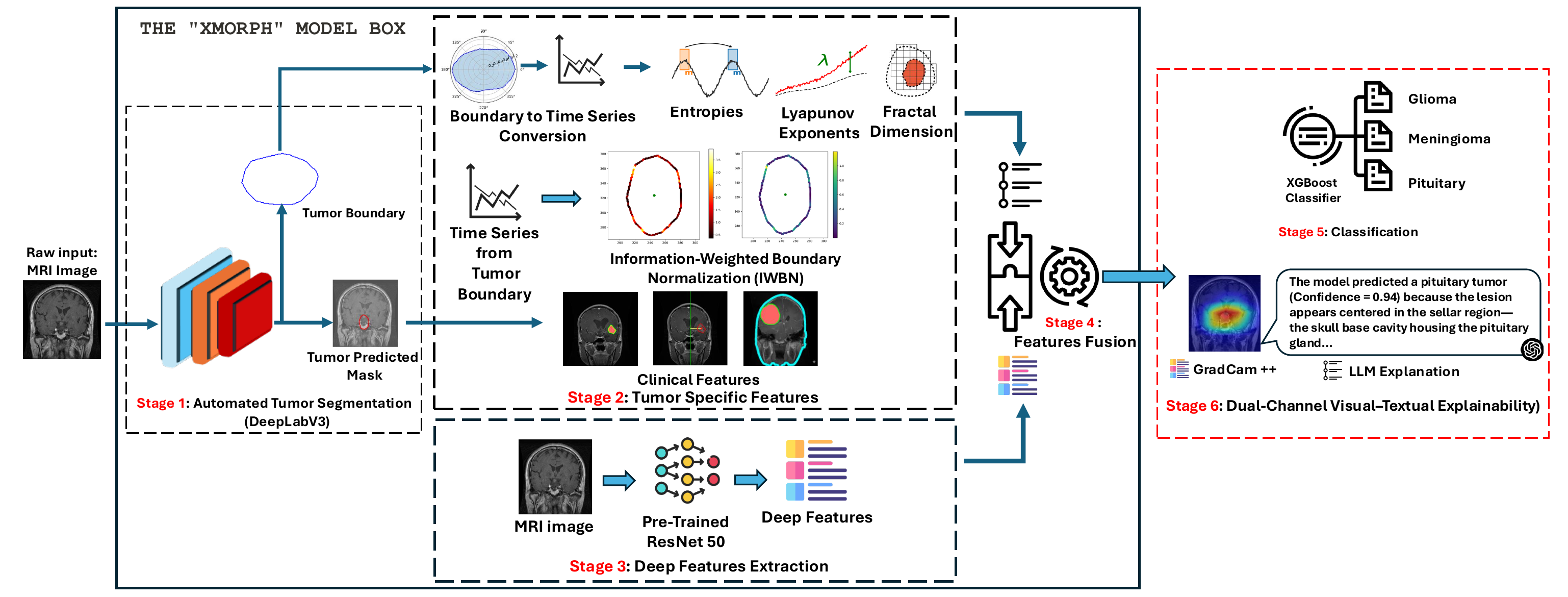}
  \caption{The end-to-end pipeline of the proposed \textbf{XMorph} framework for explainable brain tumor classification, illustrating the six-stage workflow.
\textbf{Stage 1 (Automated Tumor Segmentation):} An input MRI is processed by a DeepLabV3 model to generate a tumor mask and extract the ROI (tumor boundary).
\textbf{Stage 2 (Tumor Specific Features Extraction):} A multi-faceted feature set is extracted. The tumor boundary is converted to a time series to compute nonlinear chaotic descriptors (e.g., Entropies, Fractal Dimension), features from our novel Information-Weighted Boundary Normalization (IWBN), and quantitative clinical biomarkers.
\textbf{Stage 3 (Deep Features Extraction):} In a parallel stream, a pre-trained ResNet-50 extracts high-level textural features from the input MRI.
\textbf{Stage 4 (Features Fusion):} The Tumor Specific and deep feature sets are concatenated into a single feature vector.
\textbf{Stage 5 (Classification):} An XGBoost classifier uses the fused vector to predict the tumor class (Glioma, Meningioma, or Pituitary).
\textbf{Stage 6 (Dual-Channel Explainability):} The framework provides two complementary explanations for transparency. A visual GradCAM++ heatmap, generated from the deep feature model, shows \textit{where} the model focuses. A textual rationale from an LLM explains \textit{why} the final XGBoost classifier made its decision by interpreting the most influential features from the fused set, including the novel chaotic and clinical descriptors.}
  \label{pipeline}
\end{figure*}

Our proposed framework, XMorph, employs a multi-stage pipeline to accurately classify brain tumors (Glioma, Meningioma, and Pituitary) while providing transparent, clinically relevant explanations. As illustrated in Figure~\ref{pipeline}, the methodology consists of six primary stages that synergistically integrate deep learning with nonlinear dynamic analysis:

\begin{itemize}
    \item \textbf{Stage 1: Automated Tumor Segmentation:}  
    An input MRI is first processed using a DeepLabV3 model to automatically localize the tumor and extract its precise Region of Interest (ROI).

    \item \textbf{Stage 2: Chaotic Feature Extraction:}  
    The tumor boundary is converted into a 1D signal to extract nonlinear chaotic features that quantify morphological complexity. This stage is enhanced by our novel \textbf{Information-Weighted Boundary Normalization (IWBN)} method, which amplifies diagnostically significant boundary irregularities.
\item \textbf{Stage 3: Deep Feature Extraction:}  
    In a parallel stream, the MRI brain image is fed into a pre-trained ResNet-50 to learn high-level textural and contextual representations that are robust to variations in appearance.

    \item \textbf{Stage 4: Hybrid Feature Fusion:}  
    The outputs from the complementary streams---chaotic features, deep features, and quantitative \textbf{clinical biomarkers} (e.g., Ring Enhancement Index, Midline Shift)---are concatenated into a single, comprehensive diagnostic signature.

    \item \textbf{Stage 5: Tumor Classification:}  
    A computationally efficient XGBoost classifier predicts the final tumor class based on the fused feature vector.

    \item \textbf{{Stage 6: Dual-Channel Visual-Textual Explainability:}} 
    Finally, a dual-channel XAI module generates both a \textit{visual explanation} (via GradCAM++) to highlight influential image regions and a \textit{textual explanation} (via an LLM) to articulate the model's diagnostic reasoning in clear, human-readable language.
   
\end{itemize}

Each of these stages is detailed in the subsequent subsections.

\subsection{Stage 1: Automated Tumor Segmentation}
The goal of this stage is to accurately isolate the tumor region from the input MRI, producing a reliable ROI for downstream feature extraction. High-quality segmentation is essential because boundary morphology, chaotic descriptors, and clinical biomarkers all depend on precise delineation of the tumor contour.

To achieve this, we employ the DeepLabV3 architecture with a ResNet-50 backbone \cite{b18}, adapted specifically for medical image segmentation. The standard DeepLabV3 output layer, originally designed for 21 semantic classes in PASCAL VOC, is replaced with a binary tumor–background classifier to focus the network exclusively on pathological region identification.

To improve robustness under class imbalance—where tumor pixels constitute only a small fraction of the image—we use a hybrid loss function combining cross-entropy and Dice loss:
\begin{equation}
    L_{\text{combined}} = L_{\text{CE}} + L_{\text{Dice}}.
\end{equation}
Cross-entropy ensures pixel-level classification consistency, while Dice loss directly optimizes the overlap between the predicted mask and ground truth mask. Dice loss is particularly effective in cases of imbalance, as it measures the ratio of the intersection over the union of the two regions, preventing the model from being biased toward predicting the dominant background class.

To enhance generalization, we apply medical-specific data augmentation, including controlled rotations ($\pm 15^{\circ}$), horizontal flips, and intensity variations. The $15^{\circ}$ rotation limit was selected to reflect plausible patient head-orientation differences during MRI acquisition without introducing unrealistic anatomical distortions.

For model selection, we prioritize the Dice coefficient during validation, as Dice directly reflects segmentation quality by quantifying how much of the predicted mask overlaps with the annotated tumor. This criterion is more clinically meaningful than validation loss alone, since accurate boundary extraction is critical for subsequent morphological and chaotic feature computation.

This segmentation stage is not intended to introduce architectural novelty; rather, it provides a reliable and efficient foundation for extracting boundary-dependent descriptors, which are central to our contribution. We evaluated both DeepLabV3 and U-Net, with U-Net achieving a Dice score of $89.72\% \pm 0.146$, while DeepLabV3 delivered higher and more stable boundary accuracy. For this reason, model selection prioritized the Dice coefficient—an indicator of mask–ground-truth overlap—over validation loss, ensuring clinically meaningful segmentation quality.

\subsection{Stage 2: Tumor Specific Features Extraction}
\label{sec:combined}
The goal of this stage is to capture a rich, interpretable signature of the tumor's morphology and clinical presentation. Drawing on evidence that malignant tumor growth exhibits chaotic behavior \cite{b6,b21}, our pipeline extracts features from three complementary sources. 

First, the 2D tumor boundary is converted into a 1D radial signal to compute a suite of morphological and chaotic features. This includes geometric metrics (e.g., Irregularity), nonlinear indices (e.g., Fractal Dimension), and advanced descriptors from our novel Information-Weighted Boundary Normalization (IWBN) method. Second, we extract quantitative clinical biomarkers from the MRI, such as the Ring Enhancement Index (REI) and Midline Shift (MLS), to ground the analysis in radiological practice. 

This multi-faceted approach yields a comprehensive feature set that quantifies the tumor's shape, complexity, and clinical characteristics, providing crucial information that is complementary to the deep features learned in the next stage.

\subsubsection{Boundary-to-Signal Conversion}
For each predicted tumor mask, we extract the primary 2D boundary contour and resample it to a fixed arc-length of $N = 256$ points to ensure consistent spatial resolution across all tumors. Let $(x_i, y_i)$ denote the coordinates of the $i$-th resampled boundary point, and let $(c_x, c_y)$ represent the centroid of the tumor mask. The radial distance from the centroid to each boundary point is computed as:
\begin{equation}
    r_i = \sqrt{(x_i - c_x)^2 + (y_i - c_y)^2}, \quad i = 1,\dots,N,
\end{equation}
where $r_i$ represents the radius at position $i$ along the boundary.

To make the representation \emph{scale-invariant}—that is, independent of the absolute tumor size—we normalize each radius by the mean radius $\bar{r} = \frac{1}{N}\sum_{i=1}^N r_i$. This yields the standard radial signal:
\begin{equation}
    S_{\text{std}, i} = \frac{r_i}{\bar{r}},
\end{equation}
where $S_{\text{std},i} > 1$ indicates a protrusion relative to the average radius, and $S_{\text{std},i} < 1$ indicates an indentation. By removing the effect of tumor scale, this normalized signal captures only the \emph{shape variability} of the boundary, forming the foundation for our geometric and nonlinear feature analysis.

\subsubsection{Geometric and Statistical Feature Extraction}
First, we extract a set of interpretable features that describe the fundamental geometry and statistical properties of the tumor boundary.
\begin{itemize} 
\item \textbf{Irregularity Index:} This is our primary measure of global boundary fluctuation, defined as the standard deviation of the normalized radial distance signal ($\sigma(S_{\text{std}})$).
\textit{Rationale:} A perfectly circular tumor would have an Irregularity Index of zero. A higher value indicates a greater deviation from a simple, regular shape, which is a key hallmark of infiltrative tumor growth.

\item \textbf{Roughness Index:} This feature quantifies the fine-scale jaggedness of the boundary by summing the absolute differences between consecutive points: $\sum |S_{\text{std}, i+1} - S_{\text{std}, i}|$.
\textit{Rationale:} Unlike the Irregularity Index, which captures large-scale fluctuations, the Roughness Index is sensitive to small, high-frequency oscillations like serrations or microlobulations, providing a complementary view of the boundary texture.
\end{itemize}

\subsubsection{Information-Weighted Boundary Normalization (IWBN)}
To capture local morphological irregularities beyond global descriptors, we introduce the  IWBN framework. While prior works often characterize tumor boundaries using single, global metrics like fractal dimension \cite{b15}, these methods do not capture the localized variations in complexity that are often hallmarks of malignancy. Our IWBN approach is, to the best of our knowledge, the first to compute a local entropy value at each point along the boundary and use this information to create a new, weighted signal. This method enhances the boundary representation by amplifying regions of high structural complexity, thereby yielding a more discriminative descriptor for classification.

\paragraph{Process Overview}
For each boundary point $i$, we compute a local entropy value $E_i$ representing the variability of its neighborhood (e.g., curvature or radial distance variance). From these, we derive normalized information weights:
\begin{equation}
w_i = \frac{0.1 + \lambda \cdot \hat{E}_i}{\frac{1}{N}\sum_{j=1}^{N}(0.1 + \lambda \cdot \hat{E}_j)}, 
\quad 
\hat{E}_i = \frac{E_i - E_{\min}}{E_{\max} - E_{\min}}
\end{equation}
where $\lambda$ controls the weighting strength. 
These weights modulate the standard normalized distances $S_{\text{std},i}$ to form the enhanced signal:
\begin{equation}
S_{\text{iw},i} = \frac{S_{\text{std},i} \cdot w_i}{\frac{1}{N}\sum_{j=1}^{N} S_{\text{std},j} \cdot w_j}
\end{equation}
thus emphasizing boundary regions with higher information content.

\paragraph{Derived Indices}
From IWBN, we define three key quantitative indices:

\begin{itemize}
    \item \textbf{Mean Local Entropy:}
    \begin{equation}
        \bar{E} = \frac{1}{N} \sum_{i=1}^{N} E_i
    \end{equation}
    where \(E_i\) is the local entropy at boundary point \(i\), and \(N\) is the total number of sampled boundary points. 
    This index measures the average structural complexity along the tumor boundary, with higher values indicating more globally irregular contours.

       \item \textbf{Weight Range:}
    \begin{equation}
        \Delta w = \max(w_i) - \min(w_i)
    \end{equation}
    where \(w_i\) is the information weight assigned to boundary point \(i\). 
    This index measures how widely the weights vary along the boundary; a larger \(\Delta w\) indicates greater heterogeneity in local information emphasis, meaning some regions contain substantially more structural complexity than others.

       \item \textbf{Enhancement Factor:}
    \begin{equation}
        \text{EF} = \frac{\sigma(S_{\text{iw}})}{\sigma(S_{\text{std}})}
    \end{equation}
    where \(\sigma(S_{\text{iw}})\) is the standard deviation of the information-weighted signal and \(\sigma(S_{\text{std}})\) is the standard deviation of the standard normalized radial signal. 
    This ratio quantifies how strongly IWBN increases the variability of the boundary representation. 
    Values \(\text{EF} > 1.0\) indicate that IWBN successfully amplifies diagnostically relevant shape irregularities, enhancing the discriminative power of the representation for downstream classification.
\end{itemize}

\subsubsection{Nonlinear and Chaotic Feature Descriptors}
To capture complex and self-organizing boundary behaviors beyond geometric shape, we extract nonlinear descriptors that quantify fractality, randomness, and chaos. The Fractal Dimension, estimated via box-counting as
\begin{equation}
D = \lim_{\epsilon \to 0} \frac{\log N(\epsilon)}{\log (1/\epsilon)},
\end{equation}
measures the self-similar complexity and space-filling capacity of the contour, where higher $D$ values correspond to more irregular, infiltrative growth. Entropy-based measures (Approximate, Sample, and Permutation Entropy) evaluate the unpredictability of boundary fluctuations, while the Largest Lyapunov Exponent:
\begin{equation}
\lambda = \lim_{t \to \infty} \frac{1}{t}\ln\frac{d(t)}{d(0)}
\end{equation}
quantifies sensitivity to initial conditions, with positive $\lambda$ indicating chaotic divergence. Together, these metrics provide a compact yet expressive characterization of nonlinear morphological complexity associated with malignant progression.

\subsubsection{Clinical Feature Extraction}
Alongside chaotic boundary descriptors, we integrated three radiological biomarkers commonly evaluated in neuro-oncology \cite{b19}, each implemented in our pipeline to capture vascularity, location, and mass effect.

\begin{itemize}
    \item \textbf{Ring Enhancement Index (REI)} quantifies peripheral contrast uptake relative to the core:
    \begin{equation}
        \text{REI} = \frac{\mu_{\text{ring}} - \mu_{\text{core}}}{\mu_{\text{core}} + \epsilon},
    \end{equation}
    where $\mu_{\text{ring}}$ and $\mu_{\text{core}}$ are mean intensities of the enhancing rim and core.

    \item \textbf{Skull-to-Tumor Distance (dskull)} measures the minimum distance between the tumor boundary and the inner skull surface:
    \begin{equation}
        d_{\text{skull}} = \min_{p \in \partial \mathcal{T}} \| p - \partial \mathcal{S} \|_2.
    \end{equation}

    \item \textbf{Midline Shift (MLS)} quantifies displacement of the brain’s anatomical midline:
    \begin{equation}
        \text{MLS} = \frac{|x_{\text{midline}} - x_{\text{falx}}|}{\text{brain width}} \times 100\%.
    \end{equation}
\end{itemize}

All three metrics were derived directly from MRI masks and intensity data, quantifying vascularity (REI), anatomical location (skull distance), and mass effect (midline shift). The midline shift was computed only on axial scans, while REI and skull distance were measured across all orientations. These features complement the boundary-based analysis by linking morphological irregularity to clinically interpretable imaging biomarkers.

\subsection{Stage 3: Deep Feature Extraction}
\label{sec:deep_features}
We use a pre-trained ResNet-50 model as our deep feature extractor. The full MRI slice---rather than the cropped ROI---is resized to $224 \times 224$ and passed through the network. Features are taken from the global average pooling layer, producing a 2048-dimensional vector. To reduce redundancy and maintain computational efficiency, we apply Principal Component Analysis (PCA) to obtain a compact deep feature representation.
ResNet-50 enables efficient transfer learning, providing rich semantic representations without training a deep network from scratch. Using the full MRI preserves global context that may be relevant for classification, while PCA ensures that only the most informative components contribute to the fusion stage.

\subsection{Stage 4: Hybrid Feature Fusion}
\label{sec:fusion}
The objective of this stage is to create a single, unified diagnostic signature that synergistically combines the strengths of the different feature types. This hybrid representation is designed to be more robust and discriminative than any individual feature set alone.

We construct the final feature vector by concatenating the outputs from the preceding stages. The PCA-reduced deep feature vector from Stage 3 is combined with the full set of Tumor Specific features from Stage 2 (which includes morphological, chaotic, and clinical biomarkers). The resulting fused representation is defined as:
\begin{equation}
f^{\text{fusion}} = [\,f^{\text{deep}}_{\text{PCA}} \,\|\, f^{\text{tsf}}\,],
\end{equation}
where $f^{\text{deep}}_{\text{PCA}}$ represents the deep features after PCA and $f^{\text{tsf}}$ represents the complete set of Tumor Specific features.
 
This hybrid fusion strategy is a core contribution of our framework. It enhances discriminative performance by integrating the abstract, learned visual embeddings from the CNN with the interpretable, domain-specific Tumor Specific features. This combination allows the model to make decisions based on both the tumor's internal texture (from deep features) and its external morphology and clinical context (from Tumor Specific features).

\subsection{Stage 5: Tumor Classification}
\label{sec:classification}
The final goal of the predictive pipeline is to accurately classify the brain tumor into one of three classes: glioma, meningioma, or pituitary. For classification, we employ an XGBoost classifier, an efficient gradient boosting algorithm. The classifier is trained on the fused feature vectors ($f^{\text{fusion}}$) from Stage 4. The model minimizes a regularized objective function to prevent overfitting and improve generalization. The objective function at a given step $t$ can be expressed as:
\begin{equation}
\mathcal{L}^{(t)} = \sum_{i=1}^{n} l(y_i, \hat{y}_i^{(t-1)} + f_t(x_i)) + \Omega(f_t)
\label{eq:xgboost_obj}
\end{equation}
where $l$ is the loss function (in our case, multiclass logistic loss), $y_i$ is the true label for sample $i$, $\hat{y}_i^{(t-1)}$ is the prediction from the previous $t-1$ trees, and $f_t(x_i)$ is the prediction of the new tree being added. The term $\Omega(f_t)$ is a regularization penalty on the complexity of the new tree, which helps to control the model's complexity. We evaluated our model's performance using a five-fold stratified cross-validation strategy to ensure robust and unbiased results.

\subsection{Stage 6: Multi-Modal Explainable AI (XAI) Framework}
A complementary contribution of this work is a dual-channel (XAI) framework that enhances model interpretability by integrating visual attention maps with language-based reasoning, enabling transparent and clinically meaningful explanations of the model’s decisions.

\subsubsection{Visual Explanation with GradCAM++} To provide visual context, we apply GradCAM++ to the final convolutional layer of our CNN feature extractor. This generates a class-specific heatmap that is overlaid on the input MRI, visually highlighting the pixels and regions that were most influential in the model's decision to select a particular class (e.g., Malignant).
\subsubsection{Large Language Model–Based Explainability}
While visual explainability methods such as GradCAM++ highlight influential image regions, they do not communicate the diagnostic reasoning behind the model’s output. To overcome this, we incorporated a Large Language Model (LLM)–based explainability layer that translates quantitative model evidence into concise, clinically interpretable narratives.

For each test sample $x_i$, the XGBoost model outputs a probability distribution $\hat{y}_i = f(x_i)$ across the tumor classes. Using SHAP analysis, we decompose the model’s prediction into additive feature contributions:
\begin{equation}
f(x_i) = \phi_0 + \sum_{j=1}^{M} \phi_{ij},
\label{eq:shap}
\end{equation}
where $\phi_0$ is the model’s global bias term and $\phi_{ij}$ denotes the contribution of feature $j$ to the prediction of sample $i$. The $k$ features with the highest absolute SHAP values $|\phi_{ij}|$ are selected as the primary explanatory factors:
\begin{equation}
\mathcal{F}_i = \text{Top-}k\{ |\phi_{ij}| \}_{j=1}^{M}.
\end{equation}

Each feature triplet $(f_j, \phi_{ij}, x_{ij})$, along with the predicted class $c_i$ and confidence $p_i = \max(f(x_i))$, is formatted into a structured textual prompt $\mathcal{P}_i$:
\begin{equation}
\mathcal{P}_i = \{ c_i, p_i, [(f_j, \phi_{ij}, x_{ij})]_{j \in \mathcal{F}_i} \}.
\end{equation}
The prompt $\mathcal{P}_i$ is then provided to a pre-trained LLM (GPT-5), which generates a clinical-style explanation $E_i$ that describes the key diagnostic cues supporting the predicted class:
\begin{equation}
E_i = \text{LLM}(\mathcal{P}_i).
\end{equation}

These LLM-generated narratives effectively translate quantitative model reasoning into interpretable medical language. When combined with GradCAM++ visualizations, the dual-channel XAI system provides both spatial and semantic transparency, allowing clinicians to simultaneously visualize \textit{where} and understand \textit{why} the model reached its decision.

To operationalize this process, Algorithm \ref{alg:llm_prompt} outlines the procedure for generating an LLM prompt for a specific test case. 
Given the model’s predicted class, confidence score, and top-$k$ SHAP-ranked features, the algorithm constructs a structured natural-language query that integrates both quantitative evidence (feature names, SHAP values, and raw feature magnitudes) and qualitative instruction (“Write a clinical-style explanation and note uncertainty”). 
This formatted prompt is then passed to the LLM, which produces a concise, interpretable diagnostic rationale $E_{\text{case}}$ grounded in the same features that drove the model’s decision.
\begin{algorithm}[h]
\caption{LLM Prompt Generation with Safety Constraints}
\label{alg:llm_prompt}
\begin{algorithmic}[1]
\renewcommand{\algorithmicrequire}{\textbf{Input:}}
\renewcommand{\algorithmicensure}{\textbf{Output:}}

\Require Predicted class $c$, Confidence $p$, SHAP values $\Phi$, Feature set $X$
\Ensure Clinical Explanation $E_{case}$

\State \textbf{Step 1: Feature Selection}
\State $\mathcal{F}_{top} \gets \text{Select top-}k \text{ features based on } |\Phi|$

\State \textbf{Step 2: Contextualization}
\State $P_{sys} \gets$ "You are an expert neuro-oncologist. Analyze the provided quantitative features to explain the tumor classification. Do NOT hallucinate clinical history not present in the data."
\State $P_{user} \gets$ "Diagnosis: " + $c$ + " (" + $p$ + "\%)."
\State $P_{user} \gets P_{user} +$ " Key Features: "

\For{each feature $f_i$ in $\mathcal{F}_{top}$}
    \State $val \gets$ GetValue($f_i$)
    \State $contrib \gets$ GetSHAP($f_i$)
    \State $P_{user} \gets P_{user} + f_i.\text{name} + ": " + val + " (Impact: " + contrib + "); "$
\EndFor

\State \textbf{Step 3: Generation}
\State $E_{case} \gets \text{GPT-5-API}(system=P_{sys}, user=P_{user})$
\State \Return $E_{case}$
\end{algorithmic}
\end{algorithm}

\section{Experimental Results}
This section evaluates X\mbox{-}Morph AI in terms of (i) tumor segmentation quality, (ii) boundary-based and clinical feature behavior, (iii) classification performance under different feature configurations, and (iv) the effectiveness of the proposed dual-channel XAI framework.

\subsection{Datasets}
We evaluate our framework on publicly available brain MRI datasets, such as the Figshare 2024 collection dataset\cite{b20}. These datasets provide a large number of MRI scans with different tumor types, which are essential for training and validating our three-class model.

\subsection{Segmentation Results}
We evaluated the performance of our DeepLabV3-based segmentation model using five standard metrics: Dice Score, Intersection over Union (IoU), Precision, Recall, and F1 Score. The dataset comprised 3,564 brain MRI scans: Glioma (n=1,426), Meningioma (n=708), Pituitary tumors (n=930), and non-tumor cases (n=500). To improve model specificity and prevent false positives on healthy brain scans, we included non-tumor images with empty ground truth masks (all background pixels). This approach addresses a common limitation where models trained exclusively 
on tumor-positive cases exhibit overconfidence and hallucinate tumors in healthy tissue.

\subsubsection{Performance Metrics}
As shown in Table~\ref{tab:performance_metrics}, the model achieved an overall Dice score of $0.932$, indicating strong agreement with ground truth. Performance was highest for Non-Tumor cases ($0.958 \pm 0.068$), suggesting the model distinguishes healthy tissue from tumors reliably. Among tumor types, Meningioma achieved strong results ($0.928 \pm 0.105$), while Glioma remained the most challenging ($0.913 \pm 0.130$) due to their infiltrative margins. Pituitary tumors performed moderately well ($0.940 \pm 0.080$). High overall precision ($0.932$) and recall ($0.940$) further demonstrate robustness and clinical relevance.

\begin{table*}[t]
    \centering
    \caption{Performance metrics for brain tumor segmentation.}
    \label{tab:performance_metrics}
    \begin{tabular}{|l|c|c|c|c|c|}
        \hline
        \textbf{Tumor Type} & \textbf{Dice Score} & \textbf{IoU Score} & \textbf{Precision} & \textbf{Recall} & \textbf{F1 Score} \\
        \hline
        Glioma & $0.913 \pm 0.130$ & $0.860 \pm 0.155$ & $0.914 \pm 0.131$ & $0.924 \pm 0.136$ & $0.913 \pm 0.130$ \\
        Meningioma & $0.928 \pm 0.105$ & $0.882 \pm 0.128$ & $0.929 \pm 0.107$ & $0.938 \pm 0.109$ & $0.928 \pm 0.105$ \\
        Pituitary & $0.940 \pm 0.080$ & $0.890 \pm 0.099$ & $0.937 \pm 0.085$ & $0.948 \pm 0.085$ & $0.940 \pm 0.080$ \\
        Non-Tumor & $0.958 \pm 0.068$ & $0.923 \pm 0.085$ & $0.959 \pm 0.076$ & $0.959 \pm 0.068$ & $0.958 \pm 0.068$ \\
        \hline
        \textbf{Overall} & $\mathbf{0.932 \pm 0.104}$ & $\mathbf{0.885 \pm 0.127}$ & $\mathbf{0.932 \pm 0.106}$ & $\mathbf{0.940 \pm 0.108}$ & $\mathbf{0.932 \pm 0.104}$ \\
        \hline
    \end{tabular}
\end{table*}

\subsubsection{Qualitative Analysis}
\begin{figure}[H]
    \centering    \includegraphics[width=\columnwidth]{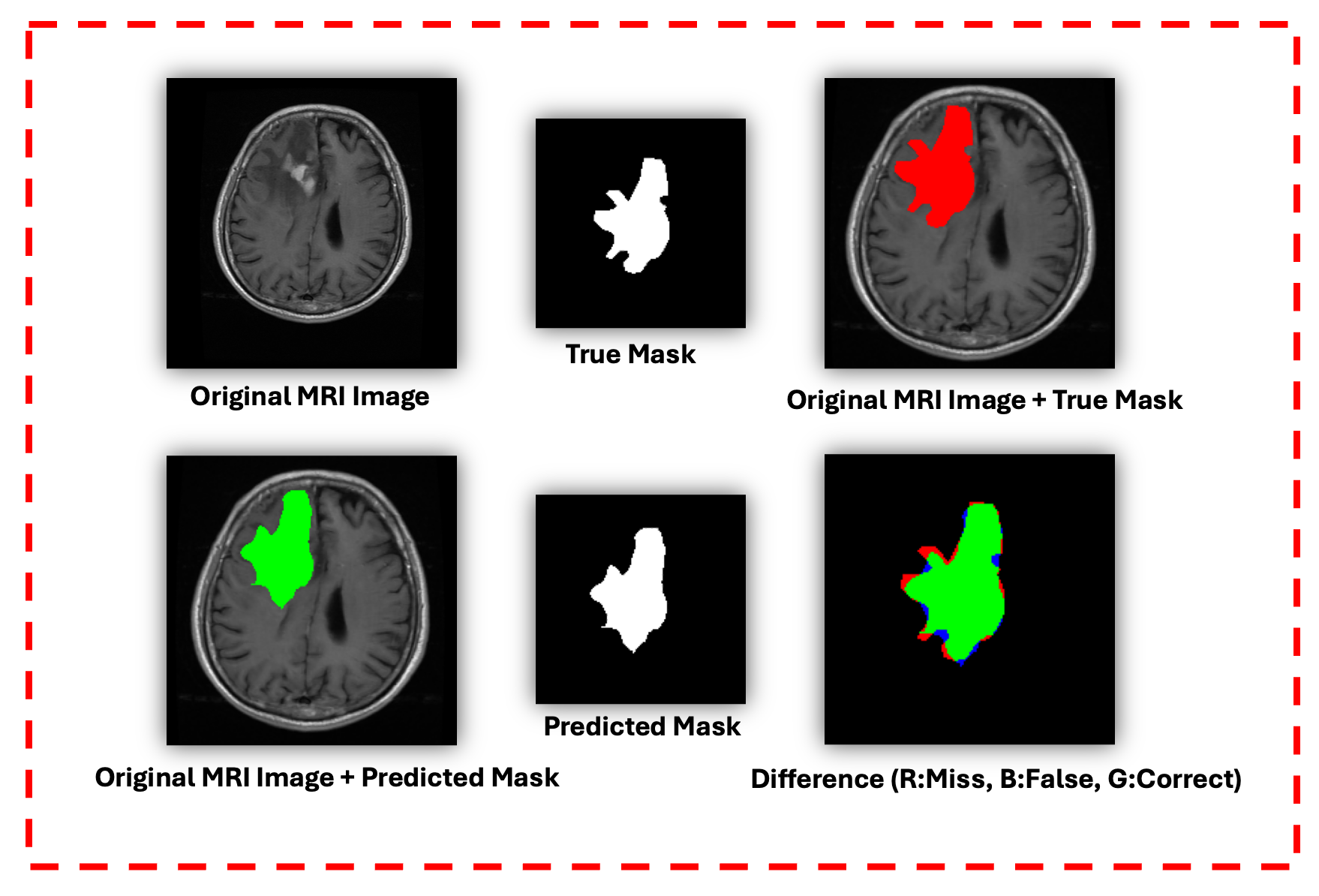}
    \caption{Example of brain tumor segmentation results showing original MRI, masks, and differences.}
    \label{fig:segmentation_results}
\end{figure}

Figure~\ref{fig:segmentation_results} illustrates a representative case (likely a glioma) with irregular and infiltrative morphology. The predicted mask (green) shows strong overlap with the ground truth (red), capturing complex tumor boundaries. The difference map highlights minimal false negatives (red) and false positives (blue), supporting the high Dice, precision, and recall scores. This visual evidence further confirms the model’s reliability for clinical tumor delineation.

\subsubsection{Segmentation Model Performance Comparison}
To evaluate the effectiveness of different segmentation architectures on brain tumor detection, we implemented and compared two state-of-the-art models: DeepLabV3 with a ResNet-50 backbone and U-Net. Both models were trained on input images with a resolution of $256 \times 256$ pixels. Training was performed for 50 epochs using the Adam optimizer with a learning rate of $1e-4$ and a weight decay of $1e-4$. The loss function combined CrossEntropy and Dice loss. To improve generalization, data augmentation techniques such as random horizontal flipping, rotation within $\pm 15^{\circ}$, and brightness/contrast adjustments were applied.
Table~\ref{tab:deeplabv3_unet} reveals that DeepLabV3 consistently outperforms U-Net across all evaluation metrics. DeepLabV3 achieved a Dice Score of 93.21\% compared to U-Net's 89.72\%, representing a 3.49 percentage point improvement. Similarly, DeepLabV3 demonstrated superior IoU performance (88.48\% vs 83.42\%) and higher precision (93.22\% vs 90.91\%).
Notably, DeepLabV3 also exhibited better consistency, as evidenced by lower standard deviations across all metrics. The standard deviation for Dice Score was $\pm 0.104$ for DeepLabV3 compared to $\pm 0.146$ for U-Net, indicating more reliable and stable predictions across different samples and tumor types.
\begin{table}[h!]
\centering
\caption{Performance comparison between DeepLabV3 and U-Net.}
\label{tab:deeplabv3_unet}
\begin{tabular}{|l|c|c|}
\hline
\textbf{Metric} & \textbf{DeepLabV3} & \textbf{U-Net} \\ \hline
Dice Score   & 93.21\% $\pm$ 0.104\% & 89.72\% $\pm$ 0.146\% \\ \hline
IoU Score    & 88.50\% $\pm$ 0.126\% & 83.42\% $\pm$ 0.160\% \\ \hline
Precision    & 93.22\% $\pm$ 0.106\% & 90.91\% $\pm$ 0.145\% \\ \hline
Recall       & 94.00\% $\pm$ 0.108\% & 89.98\% $\pm$ 0.157\% \\ \hline
F1 Score     & 93.20\% $\pm$ 0.104\% & 89.71\% $\pm$ 0.146\% \\ \hline
\end{tabular}
\end{table}

\begin{figure*}[h!]
  \includegraphics[width=1.0\textwidth]{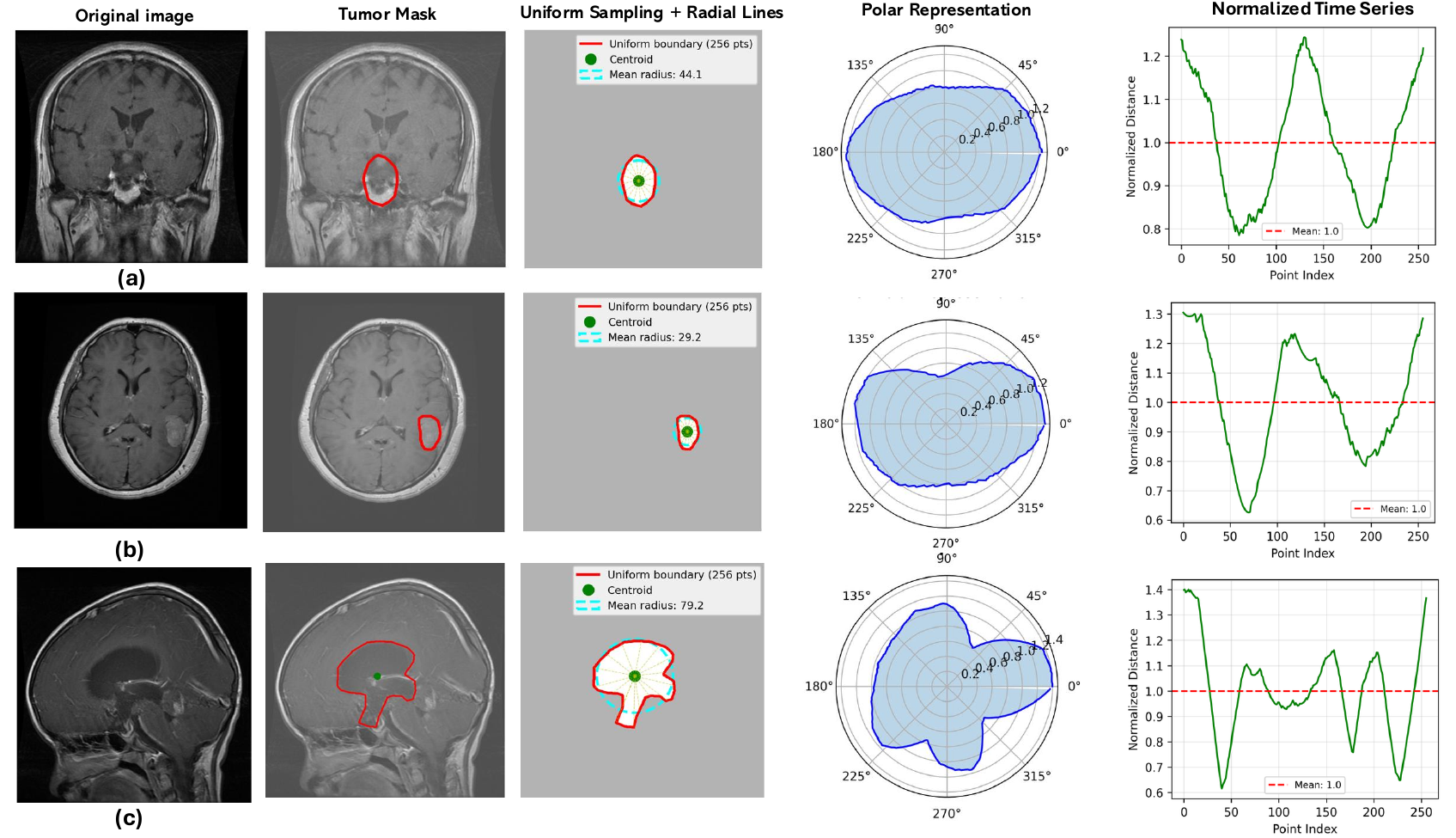}
  \caption{\textbf{The boundary-to-signal pipeline reveals distinct morphological signatures across different tumor types.} The pipeline converts 2D tumor boundaries (Column 2) into 1D normalized time-series signals (Column 5). Each row illustrates a representative clinical case, demonstrating the direct relationship between visual morphology and the resulting signal characteristics.
  \textbf{(a) Pituitary Tumor:} A visually regular, well-circumscribed boundary translates into a smooth, low-amplitude signal, quantitatively confirmed by a low Irregularity Index (STD = 0.142).
  \textbf{(b) Meningioma:} A characteristic lobulated boundary produces a signal with more pronounced variations, reflecting an intermediate structural complexity (Irregularity Index = 0.157).
  \textbf{(c) Glioma:} A highly infiltrative and poorly defined malignant boundary generates a chaotic, non-periodic signal with sharp, high-amplitude oscillations, corresponding to a significantly higher Irregularity Index (STD = 0.253).
  This clear monotonic progression from a smooth to a chaotic signal provides strong visual evidence that the pipeline robustly captures clinically relevant morphological differences.}
\label{fig:signals}
\end{figure*} 
\begin{figure*}[h!]
  \centering
  \includegraphics[width=0.8\textwidth]{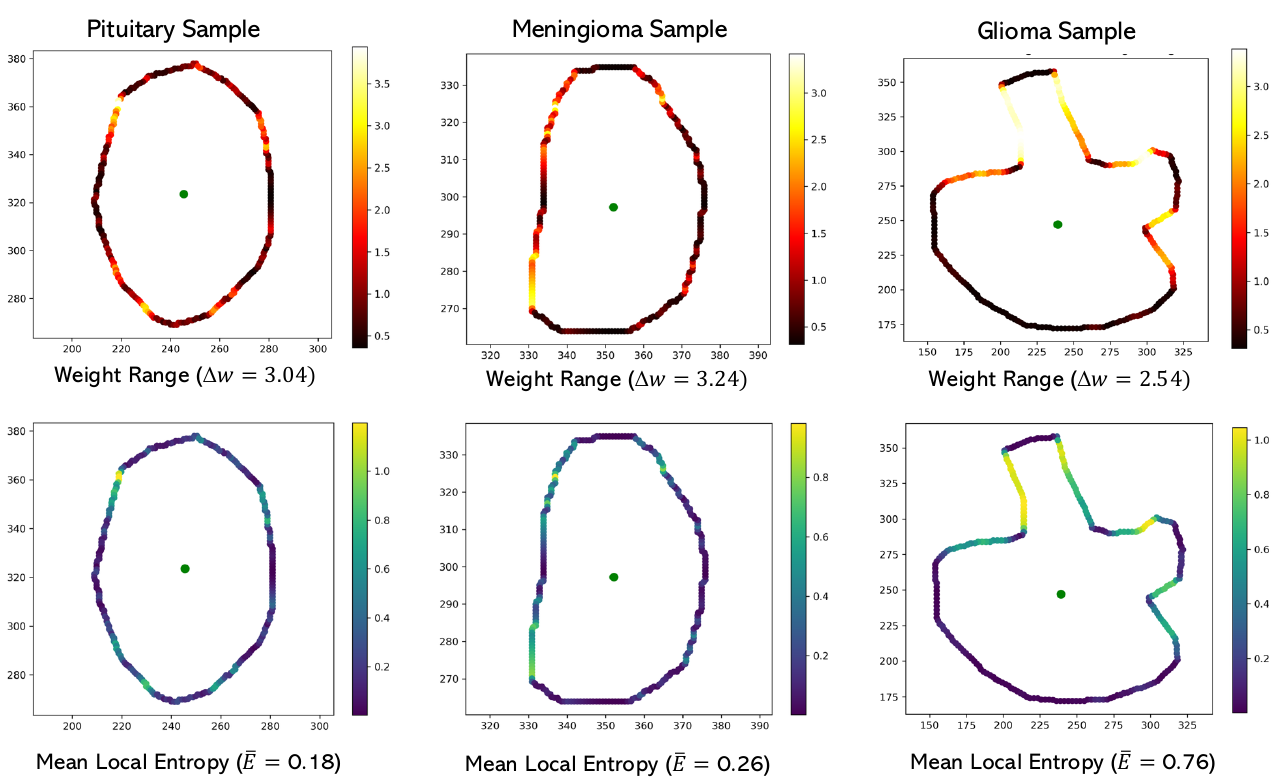} 
  \caption{\textbf{Visual Validation of IWBN across Tumor Types.} This figure illustrates how our novel IWBN method identifies and emphasizes regions of high morphological complexity. Each column represents a different tumor class, with the top row showing the final Information Weights and the bottom row showing the underlying Local Entropy that generated them.
  \textbf{Top Row (Information Weights):} Visualizes the amplification factor applied to each point on the boundary. Hotter colors (red/yellow) indicate higher weights, signifying regions of greater diagnostic interest.
  \textbf{Bottom Row (Local Entropy):} Maps the intrinsic complexity of the boundary. Brighter colors (yellow) correspond to segments with higher irregularity and unpredictability.
  \textbf{Pituitary:} The benign, regular boundary exhibits uniformly low entropy, resulting in low and evenly distributed weights. The method correctly identifies this as a simple structure.
  \textbf{Meningioma:} The method accurately pinpoints the lobulated portion of the boundary as a localized region of high entropy, assigning it a correspondingly high weight.
  \textbf{Glioma:} The malignant, infiltrative boundary shows widespread, high-magnitude entropy. The IWBN method correctly assigns high weights to these numerous chaotic segments, effectively highlighting the features of malignancy.}
\label{fig:iwbn_results}
\end{figure*}

\subsection{Boundary Morphology and Signal Characteristics}
To validate our feature engineering approach, we first qualitatively analyzed the relationship between the known clinical morphology of different tumor types and their corresponding 1D boundary signals. Figure~\ref{fig:signals} illustrates our complete boundary-to-signal pipeline for three representative cases, revealing distinct and class-specific morphological signatures.

The pipeline successfully transforms the complex 2D tumor boundary into a quantitative, size-invariant time series. The visual and quantitative differences between the tumor classes are immediately apparent:

\begin{itemize}
    \item \textbf{Pituitary Tumor (Figure~\ref{fig:signals}a):} As a typically benign and well-circumscribed lesion, the pituitary tumor exhibits a highly regular, ovoid shape. This translates into a smooth, periodic time series with low-amplitude oscillations. The resulting Irregularity Index is correspondingly low, quantitatively confirming the boundary's high degree of regularity.

    \item \textbf{Meningioma (Figure~\ref{fig:signals}b):} This meningioma, while also benign, presents with a characteristic lobulated morphology. These large-scale indentations and protrusions are accurately captured by the pipeline, producing a time series with more pronounced, yet still relatively smooth, variations. The Irregularity Index is moderately higher, reflecting this intermediate level of structural complexity.

    \item \textbf{Glioma (Figure~\ref{fig:signals}c):} In stark contrast, the malignant glioma displays a highly infiltrative and poorly defined boundary. This aggressive growth pattern generates a chaotic and non-periodic time series characterized by sharp, high-amplitude, and unpredictable oscillations. This is quantitatively reflected in a significantly higher Irregularity Index, which captures the extreme irregularity of the malignant contour.
\end{itemize}

The clear monotonic progression observed—from the smooth signal of the benign pituitary tumor to the chaotic signal of the malignant glioma—provides strong initial evidence that our 1D signal representation robustly captures clinically relevant morphological differences. This demonstrates that boundary irregularity, as quantified by our pipeline, is a powerful biomarker for distinguishing between tumor types, thereby motivating the quantitative analysis across the entire dataset presented in the following section.

\subsection{Validation of Information-Weighted Normalization}

Having established the baseline morphological signatures, we now evaluate the effectiveness of our novel IWBN. The objective of IWBN is to selectively amplify boundary signal irregularities in diagnostically relevant regions of high local complexity. Figure~\ref{fig:iwbn_results} illustrates this effect across representative tumor cases, while quantitative values further confirm the interpretability of the approach.

\begin{itemize}
    \item For the \textbf{Pituitary tumor}, the boundary exhibits globally low complexity. Both the Mean Local Entropy (0.19) and Weight Range (3.14) are small, indicating a smooth, homogeneous contour. The IWBN method thus correctly identifies this tumor as compact and well-circumscribed.
    \item For the \textbf{Meningioma}, the Mean Local Entropy is modest (0.29), yet the Weight Range is relatively high (3.39). This pattern reveals that the tumor boundary is largely regular but contains localized lobulations, which are effectively highlighted by IWBN as focal sources of complexity.
    \item For the \textbf{Glioma}, the Mean Local Entropy (0.76) and its variability (0.67) are markedly higher than in the other two tumor types, while the Weight Range (2.99) indicates widespread but uneven irregularities. IWBN generates multiple high-weight “hot spots,” capturing the diffuse, chaotic protrusions characteristic of infiltrative malignancy.
\end{itemize}

The quantitative trends align with known tumor biology: gliomas are globally chaotic, meningiomas are mostly smooth with localized irregularities, and pituitary adenomas are highly regular. 

Most importantly, the Enhancement Factor defined as the ratio of the standard deviation of the IWBN signal to that of the standard normalized signal provides a direct measure of irregularity amplification. Higher values indicate stronger emphasis of irregular features. In our representative cases, the glioma exhibited the strongest enhancement, confirming that IWBN is most sensitive to malignant boundary morphology. This demonstrates that IWBN not only preserves clinically relevant shape information but also magnifies subtle irregularities, thereby yielding a more discriminative feature space for classification.
\subsection{Visualization of Clinical Features Across Tumor Types}
\begin{figure}[h!]
  \centering
  \includegraphics[width=1.0\linewidth]{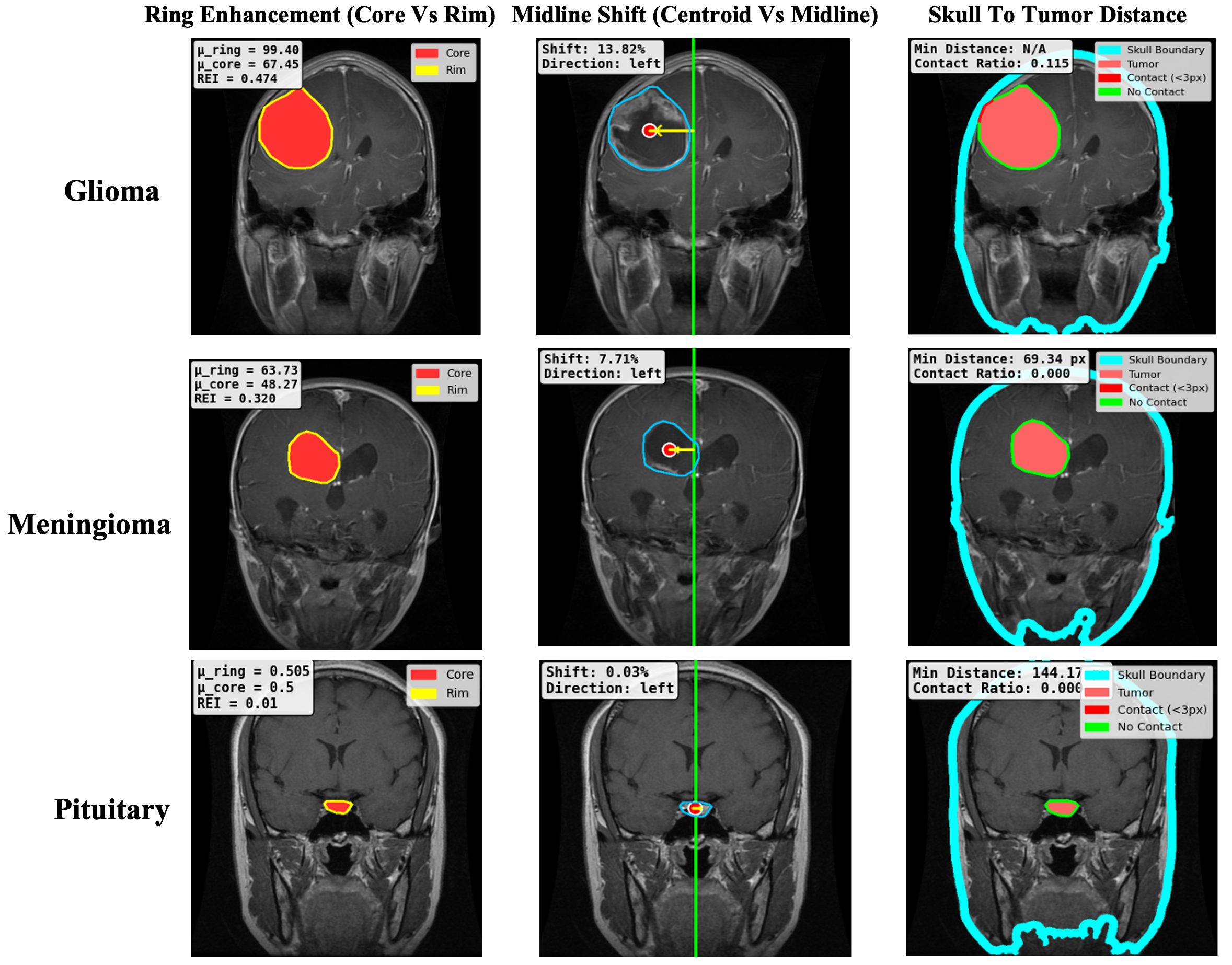}
  \caption{Visualization of three key radiological biomarkers across tumor types:
    \textbf{(Left)} Ring Enhancement Index (core vs.\ rim intensity), 
    \textbf{(Center)} Midline Shift (centroid deviation from the anatomical midline), and 
    \textbf{(Right)} Skull-to-Tumor Distance (minimum distance and contact ratio). 
    The figure demonstrates increasing enhancement irregularity and deformation gradient from pituitary $\rightarrow$ meningioma $\rightarrow$ glioma, reflecting both vascular and spatial indicators of malignancy.}
  \label{fig:clinical_features}
\end{figure}

Figure~\ref{fig:clinical_features} illustrates three clinically interpretable radiological biomarkers—(REI), Midline Shift, and Skull-to-Tumor Distance—for representative cases of glioma, meningioma, and pituitary adenoma.
For the glioma, the high REI of 0.47\% indicates strong peripheral enhancement around a necrotic core, a characteristic of infiltrative, high-grade lesions. This is accompanied by a pronounced mass effect, evidenced by a significant midline shift of 13.82\%. As the tumor has expanded to the brain's surface, its partial skull contact (contact ratio $\approx 0.12$) is also apparent.

In the provided meningioma example, the lesion exhibits a lower REI (0.32), suggesting more uniform enhancement. Unlike a typical extra-axial tumor, this particular case shows a lesion located deep within the brain parenchyma. This deep-seated position is quantified by a large minimum skull-to-tumor distance of 69.34 px and a contact ratio of zero. The mass effect results in a moderate midline shift of 7.71\%.

In contrast, the pituitary adenoma demonstrates features consistent with its typical location. A minimal REI (0.01) confirms its homogeneous enhancement without a necrotic rim. Its origin in the sella turcica, at the center of the skull base, logically results in a negligible midline shift (0.03\%) and the largest measured skull-to-tumor distance (144.17 px).

To further validate these observations, we computed class-level statistics of all extracted geometric and irregularity-based features across glioma, meningioma, and pituitary tumors. The aggregated results (Table~\ref{tab:all_features}) highlight consistent differences that align with known clinical behavior of benign and malignant tumors.
\begin{table*}[ht]
\centering
\caption{Summary of morphological and information-weighted features across tumor classes. 
Higher roughness and irregularity in gliomas reflect diffuse infiltration, while higher enhancement and weight range in meningiomas indicate stronger local heterogeneity captured by the IWBN method.}
\label{tab:all_features}
\begin{tabular}{|l|c|c|c|c|c|c|c|}
\hline
\textbf{Class} & \textbf{Irregularity} & \textbf{Roughness} & \textbf{Area} & \textbf{Mean Radius} & \textbf{Mean Local Entropy} & \textbf{Weight Range} & \textbf{Enh. Factor} \\ \hline
Glioma     & 0.1801 $\pm$ 0.0936 & 3.001 $\pm$ 1.0483 & 5778.9 $\pm$ 4044.0 & 41.4 $\pm$ 15.4 & 0.7563 $\pm$ 0.3399 & 2.99 $\pm$ 0.65 & 5.28 $\pm$ 2.87 \\ \hline
Meningioma & 0.1215 $\pm$ 0.0632 & 2.443 $\pm$ 0.6120 & 4650.0 $\pm$ 3128.7 & 36.7 $\pm$ 12.9 & 0.2899 $\pm$ 0.2939 & 3.39 $\pm$ 0.58 & 7.19 $\pm$ 3.25 \\ \hline
Pituitary  & 0.1435 $\pm$ 0.0738 & 2.750 $\pm$ 0.6749 & 2163.9 $\pm$ 1633.0 & 25.1 $\pm$ 8.8 & 0.1901 $\pm$ 0.2301 & 3.14 $\pm$ 0.53 & 6.19 $\pm$ 3.25 \\ \hline
\end{tabular}
\end{table*}

This clear separation between malignant and benign distance signals demonstrates that nonlinear complexity measures provide robust, discriminative features for classification of brain tumors.

\section{Classification Performance}

This section evaluates the performance of our proposed hybrid fusion strategy. We first detail the experimental setup used for a fair comparison and then present the quantitative results, including an analysis of the Receiver Operating Characteristic (ROC) curves.
\subsection{Experimental Setup}
To validate the effectiveness of our hybrid approach, we conducted a comparative analysis of three distinct feature configurations:
\begin{enumerate}
    \item \textbf{Tumor Specific Features Only:} Using the full set of morphological, chaotic, and clinical biomarkers from Stage 2.
    \item \textbf{Deep Features Only:} Using the PCA-reduced ResNet-50 features from Stage 3.
    \item \textbf{Fused Hybrid Features:} Using the concatenated vector of both Tumor Specific and deep features, as described in Stage 4.
\end{enumerate}

To ensure reproducibility and fair comparison, all configurations were evaluated using the same XGBoost classifier. The deep feature extraction used a ResNet-50 pre-trained on ImageNet, implemented in PyTorch 2.0 on NVIDIA GPUs. Detailed hyperparameters for every stage of the pipeline—from segmentation to classification—are provided in Table~\ref{tab:hyperparams}.

\begin{table}[h!]
\centering
\caption{Detailed hyperparameter configuration for XMorph modules.}
\label{tab:hyperparams}
\resizebox{\columnwidth}{!}{
\begin{tabular}{|l|l|}
\hline
\textbf{Stage} & \textbf{Configuration / Parameters} \\ \hline
\textbf{Segmentation} & \begin{tabular}[c]{@{}l@{}}DeepLabV3 (ResNet-50), Input $256 \times 256$, Batch 16\\ Optimizer: Adam ($lr=10^{-4}$), Loss: $L_{CE} + L_{Dice}$\end{tabular} \\ \hline
\textbf{Specific Features} & \begin{tabular}[c]{@{}l@{}}Boundary $N=256$, IWBN $\alpha=0.5$, REI $\epsilon=10^{-6}$\\ Entropy ($m=2, r=0.2\sigma$), Fractal Dim. (Box-counting)\end{tabular} \\ \hline
\textbf{Deep Features} & ResNet-50 (ImageNet), Global Avg Pooling, PCA (95\% var) \\ \hline
\textbf{Classification} & XGBoost: 300 estimators, Max Depth 8, Learning Rate 0.05 \\ \hline
\end{tabular}%
}
\end{table}

Feature vectors were z-score normalized within each training fold. Model performance was assessed using a five-fold stratified cross-validation, and we report the mean and standard deviation for key metrics, including accuracy, sensitivity, and specificity.

\begin{table*}[h!]
\centering
\caption{Comparison of classification performance (mean $\pm$ standard deviation) across feature configurations using five-fold cross-validation.}
\label{tab:fusion_results}
\begin{tabular}{|l|c|c|c|}
\hline
\textbf{Feature Configuration} & \textbf{Accuracy} & \textbf{Specificity} & \textbf{Sensitivity} \\ \hline
Tumor Specific Features (XGBoost) & $0.900 \,\pm\, 0.013$ & $0.950 \,\pm\, 0.007$ & $0.895 \,\pm\, 0.012$ \\ \hline
Deep Features (ResNet-50 + XGBoost) & $0.930 \,\pm\, 0.008$ & $0.962 \,\pm\, 0.005$ & $0.918 \,\pm\, 0.010$ \\ \hline
Fused Deep + Tumor Specific Features & $\mathbf{0.960 \,\pm\, 0.010}$ & $\mathbf{0.983 \,\pm\, 0.005}$ & $\mathbf{0.962 \,\pm\, 0.012}$ \\ \hline
\end{tabular}
\end{table*}

\subsection{Model Performance and ROC Analysis}
\label{sec:roc_analysis}

As shown in Table~\ref{tab:fusion_results}, Tumor Specific features alone form a strong baseline, achieving an accuracy of $0.90 \pm 0.01$ with balanced sensitivity ($0.89$) and specificity ($0.95$). These interpretable descriptors capture key aspects of tumor morphology, vascularity, and structural displacement, demonstrating their diagnostic value even without deep representations. Using deep features extracted from the pretrained ResNet-50 further improved performance to $0.93 \pm 0.008$ accuracy, with sensitivity of $0.92$ and specificity of $0.96$. This gain highlights the network’s ability to learn high-level visual and contextual cues from MRI data. The best results were obtained by fusing Tumor Specific and deep features, reaching $0.964 \pm 0.010$ accuracy, $0.962$ sensitivity, and $0.983$ specificity—confirming that the two feature types provide complementary diagnostic information.

\begin{figure}[H]
    \centering    \includegraphics[width=0.8\columnwidth]{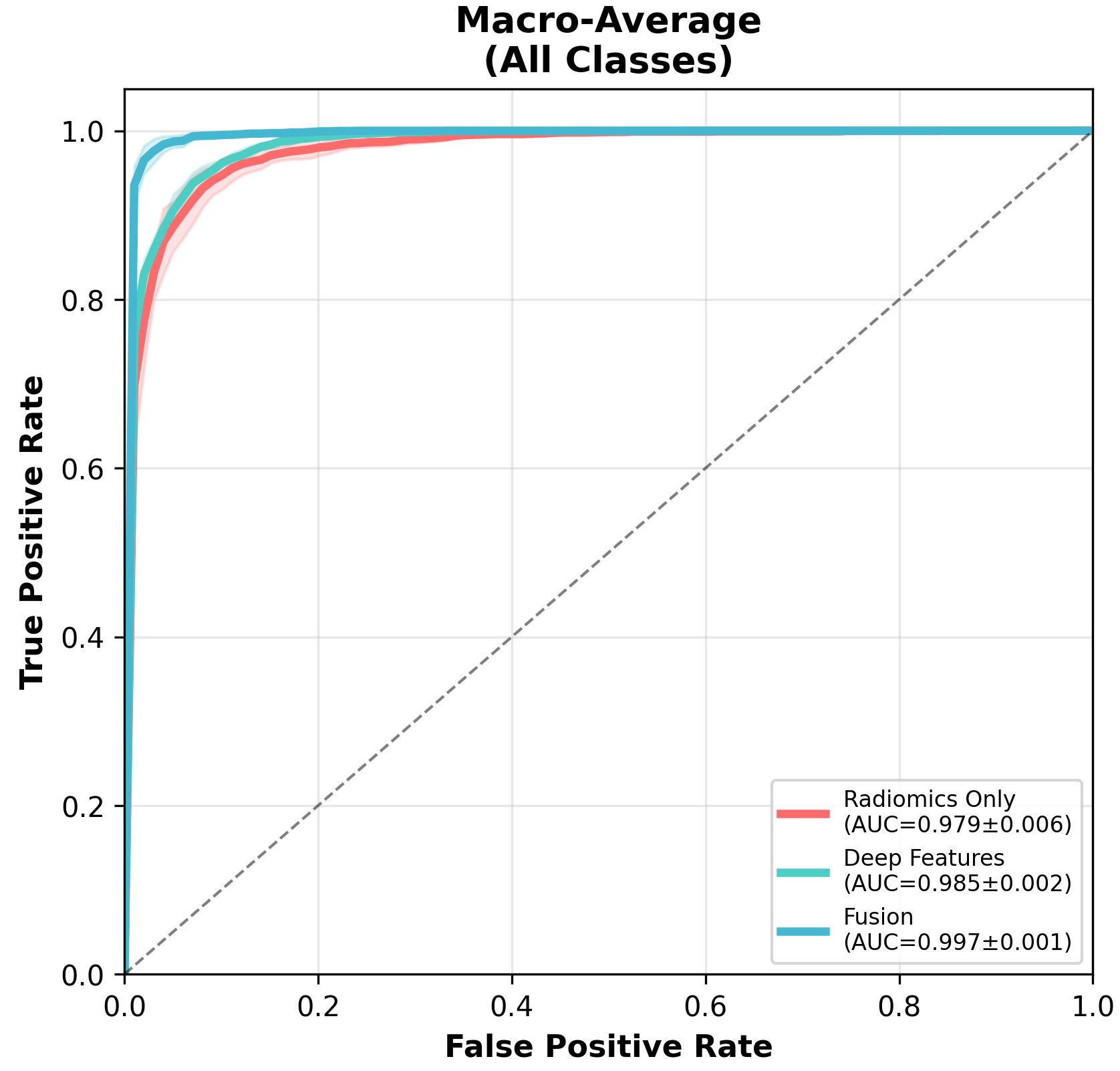}
    \caption{ROC curves comparing the performance of Tumor Specific, deep, and fused feature configurations for brain tumor classification. The fusion model demonstrates a consistently higher true positive rate across all tumor classes, indicating improved sensitivity–specificity balance. The top panels illustrate per-class ROC curves for glioma, meningioma, and pituitary tumors, while the bottom panels present macro-averaged ROC, AUC comparisons, and performance summaries.}
    \label{fig:roc_curves}
\end{figure}

For ROC analysis, as shown in Figure~\ref{fig:roc_curves}, the fusion model consistently outperformed both individual modalities across all tumor types. These findings indicate that Tumor Specific features contribute complementary, clinically interpretable information that enhances the discriminative power and generalization of deep learning models.

\begin{figure*}[t]
    \centering
    \includegraphics[width=\textwidth]{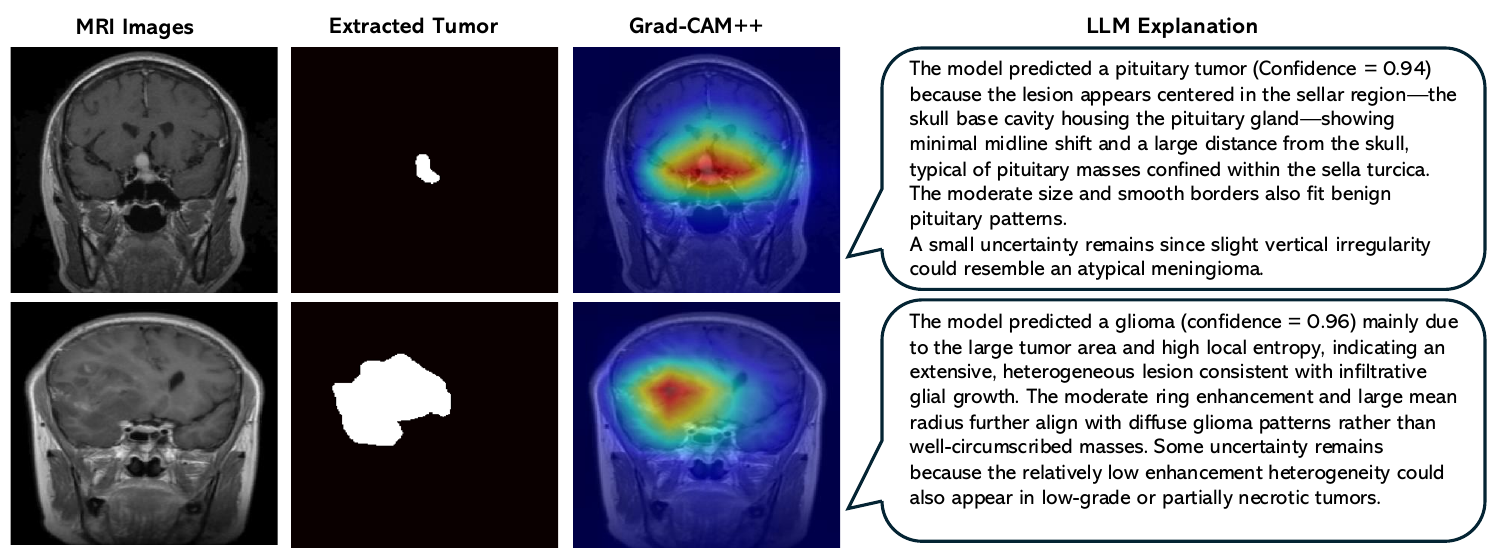}
    \caption{\textbf{Dual-channel explainability results highlighting the role of chaotic and information-weighted features.}
    Each row shows: (1) the original MRI, (2) the extracted tumor mask, (3) GradCAM++ visualization of salient regions, and (4) the LLM-generated explanation derived from fused SHAP-weighted features.
    The pituitary case (top) demonstrates low boundary entropy and minimal mass effect, while the glioma case (bottom) reflects high local entropy, large radius, and heterogeneous enhancement-patterns amplified by our chaotic feature extraction.
    Together, the spatial (GradCAM++) and semantic (LLM) channels reveal how nonlinear boundary irregularity and entropy-based cues contribute to diagnostic reasoning, translating complex quantitative patterns into human-interpretable clinical insight.}
    \label{fig:llm_explainability}
\end{figure*}

\subsection{Dual-Channel Visual–Textual Explainability}

Our framework's final stage moves beyond simple prediction to provide transparent, clinically relevant explanations. As shown in Figure~\ref{fig:llm_explainability}, we integrate GradCAM++ visual maps with our LLM-based textual reasoning module to create a cohesive dual-channel explanation. While GradCAM++ localizes spatially salient regions in the MRI, the LLM translates the model’s internal reasoning into a concise clinical narrative. A key strength of this approach is the LLM's ability to interpret and articulate the diagnostic significance of our novel Tumor Specific features. Standard XAI methods can explain which pixels are important, but they cannot explain concepts like ``high boundary entropy'' or ``infiltrative growth patterns.'' Our LLM-based module bridges this gap. By receiving the top-k SHAP-ranked features—which often include our chaotic and IWBN-derived metrics—the LLM can generate reasoning grounded in advanced morphological concepts. For example, in the glioma case (Malignant), the explanation is not just based on pixel intensity but is directly linked to quantitative evidence of malignancy, such as high local entropy, a large mean radius, and a strong ring enhancement index. These are precisely the nonlinear and clinical features our framework is designed to capture. The LLM effectively translates these numerical descriptors into a clinically intuitive rationale: the tumor is likely a glioma because it exhibits features consistent with ``diffuse, infiltrative pathology.'' In contrast, for the pituitary case (Benign), the LLM highlights features of benignancy like minimal midline shift and low boundary entropy. By jointly leveraging quantitative chaotic descriptors and semantic LLM interpretation, XMorph creates an interpretable pathway from complex numerical features to actionable clinical insight, demonstrating the unique value of combining Tumor Specific feature engineering with modern language models.

\section{Comparative Analysis}
To highlight the unique positioning of XMorph within the current landscape of brain tumor diagnostics, we conduct a comparative analysis against prominent state-of-the-art (SOTA) methodologies. A primary barrier to the clinical adoption of AI in oncology is the ``black box'' nature of deep learning. While earlier studies such as \cite{b4} and \cite{b21} achieved high accuracy, they provided no interpretability for clinical verification. Newer models like \cite{b7} and \cite{b8} introduced visual XAI via Grad-CAM heatmaps, but they lack the semantic reasoning required to articulate complex diagnostic logic. Conversely, newer LLM-based approaches like \cite{b9} generate textual reports but ignore the morphological boundary complexity (captured by metrics like FD, ApEn, and LE) that is vital for identifying infiltrative tumor growth.
As illustrated in Table~\ref{tab:capability}, XMorph is the first framework to holistically unify nonlinear boundary metrics (IWBN), quantitative clinical biomarkers (REI, MLS, dskull), and dual-channel (visual-textual) explainability. While \cite{b13} and \cite{b14} utilize fractal dimension for characterization, they do not integrate the specific chaotic measures or the radiological biomarkers required to ground the diagnosis in established medical practice. By bridging these gaps, XMorph provides a more comprehensive and clinically-grounded foundation for brain tumor classification than existing unimodal or partially explainable systems.
\begin{table}[h!]
\centering
\caption{Capability comparison of XMorph vs. state-of-the-art approaches.}
\label{tab:capability}
\renewcommand{\cmark}{\textcolor{green!70!black}{\ding{51}}}
\renewcommand{\xmark}{\textcolor{red}{\ding{55}}}
\resizebox{\columnwidth}{!}{%
\begin{tabular}{|l|c|c|c|c|c|c|c|c|}
\hline
\textbf{Feature / Capability}
& \textbf{\cite{b4}}
& \textbf{\cite{b21}}
& \textbf{\cite{b7}}
& \textbf{\cite{b8}}
& \textbf{\cite{b13}}
& \textbf{\cite{b14}}
& \textbf{\cite{b9}}
& \textbf{XMorph} \\ \hline

Deep Feature Learning
& \cmark & \cmark & \cmark & \cmark & \cmark & \cmark & \cmark & \textbf{\cmark} \\ \hline

Fractal Dimension (FD)
& \xmark & \xmark & \xmark & \xmark & \cmark & \cmark & \xmark & \textbf{\cmark} \\ \hline

Chaotic Metrics (ApEn, LE)
& \xmark & \xmark & \xmark & \xmark & \xmark & \xmark & \xmark & \textbf{\cmark} \\ \hline

IWBN (Boundary Enhancement)
& \xmark & \xmark & \xmark & \xmark & \xmark & \xmark & \xmark & \textbf{\cmark} \\ \hline

Clinical Biomarkers (REI, MLS, Dskull)
& \xmark & \xmark & \xmark & \xmark & \xmark & \xmark & \xmark & \textbf{\cmark} \\ \hline

Visual XAI (Heatmaps)
& \xmark & \xmark & \cmark & \cmark & \cmark & \xmark & \xmark & \textbf{\cmark} \\ \hline

Textual XAI (LLM Rationales)
& \xmark & \xmark & \xmark & \xmark & \xmark & \xmark & \cmark & \textbf{\cmark} \\ \hline

\end{tabular}%
}
\end{table}

\section{Conclusion}

In this work, we introduced XMorph, a hybrid deep intelligence framework for explainable three-class brain tumor classification that synergizes deep learning with nonlinear dynamics. Our primary contribution is the novel Information-Weighted Boundary Normalization (IWBN) scheme, which enhances morphological representation by selectively amplifying diagnostically relevant regions of the tumor boundary. The IWBN-derived indices, combined with other chaotic and clinical features, capture clinically consistent differences between glioma, meningioma, and pituitary tumors. Our second contribution is a dual-channel XAI module that couples GradCAM++ heatmaps with LLM-generated clinical narratives, linking spatial attention to feature-driven reasoning in a form accessible to radiologists. Experimentally, our framework demonstrated strong performance, with the segmentation backbone achieving a Dice score of 0.932 and the hybrid classification model reaching an accuracy of 96.0\%. Our results confirm that the fused representation, which combines Tumor Specific features with deep embeddings, significantly outperforms models based on either feature set alone. This finding highlights a key takeaway: IWBN-derived morphology, chaotic descriptors, and clinical biomarkers provide complementary information that enhances both the discriminative power and the interpretability of deep learning systems. Future work will extend XMorph to exploit multi-parametric MRI by jointly modeling T1, T1c, T2, and FLAIR sequences. We also plan to evaluate the framework on larger, multi-center cohorts and explore the integration of advanced Vision–Language Models (VLMs) for more deeply grounded clinical reasoning. Ultimately, our goal is to develop a fully integrated, end-to-end system that unifies multimodal imaging and language-based explainability for real-time, clinically deployable decision support, paving the way for more transparent and trustworthy medical AI.
\section*{Acknowledgments}
M.A. acknowledges a startup fund from Georgia State University (GSU), provided by the Department of Computer Science and the College of Arts and Sciences. M.A. also acknowledges a Faculty Internal Grant (FIG) from Georgia State University (GSU), and an industrial grant from Advanced Micro Devices (AMD), Inc.

\bibliographystyle{IEEEtran}
\bibliography{refs}

\vspace{12pt}
\end{document}